\documentclass[runningheads]{llncs}

\usepackage{graphicx}
\usepackage{hyperref}
\usepackage{booktabs}
\usepackage{xcolor}

\usepackage{pgfplots}
\pgfplotsset{compat=1.17}
\usepackage{amsmath}
\usepackage{color, colortbl}

\DeclareMathOperator{\opand}{and}

\newcommand{\cc}{\cellcolor{lightgray}}

\begin{document}

\title{Gender Stereotyping Impact in Facial Expression Recognition}

\author{Iris Dominguez-Catena\orcidID{0000-0002-6099-8701} \and
Daniel Paternain\orcidID{0000-0002-5845-887X} \and
Mikel Galar\orcidID{0000-0003-2865-6549}}

\authorrunning{I. Dominguez-Catena et al.}

\institute{Institute of Smart Cities (ISC), Department of Statistics, Computer Science and Mathematics\\
Public University of Navarre (UPNA)\\
Arrosadia Campus, 31006, Pamplona, Spain \\
\email{\{iris.dominguez, mikel.galar, daniel.paternain\}@unavarra.es}}

\maketitle

\begin{abstract}
Facial Expression Recognition (FER) uses images of faces to identify the emotional state of users, allowing for a closer interaction between humans and autonomous systems. Unfortunately, as the images naturally integrate some demographic information, such as apparent age, gender, and race of the subject, these systems are prone to demographic bias issues. In recent years, machine learning-based models have become the most popular approach to FER. These models require training on large datasets of facial expression images, and their generalization capabilities are strongly related to the characteristics of the dataset. In publicly available FER datasets, apparent gender representation is usually mostly balanced, but their representation in the individual label is not, embedding social stereotypes into the datasets and generating a potential for harm. Although this type of bias has been overlooked so far, it is important to understand the impact it may have in the context of FER. To do so, we use a popular FER dataset, FER+, to generate derivative datasets with different amounts of stereotypical bias by altering the gender proportions of certain labels. We then proceed to measure the discrepancy between the performance of the models trained on these datasets for the apparent gender groups. We observe a discrepancy in the recognition of certain emotions between genders of up to $29 \%$ under the worst bias conditions. Our results also suggest a safety range for stereotypical bias in a dataset that does not appear to produce stereotypical bias in the resulting model. Our findings support the need for a thorough bias analysis of public datasets in problems like FER, where a global balance of demographic representation can still hide other types of bias that harm certain demographic groups.
\end{abstract}

\section{Introduction}
The development of technology in the last decades, especially in Machine Learning (ML) and Artificial Intelligence (AI), has exposed an ever-growing portion of the population to autonomous systems. These systems, from the mundane autocorrector in mobile devices to the critical autopilot in self-driving cars, impact the lives of people around the world. Despite their continuous improvement in all respects, this impact is not always positive. A point of particular concern is when the mistakes our AI systems make systematically harm certain demographic groups. We call this behavior an unwanted bias. Unwanted biases can be based on several demographic characteristics, the most common being age, sex, and race \cite{Keyes2018,Prabhu2020,Buolamwini2018}.

These biases have been studied and classified into many types according to the stage of the ML life cycle from which they originate \cite{Suresh2021}. Although all sources of bias must be taken into account to develop fair systems, dataset bias has gained special relevance in the last decade. For many ML applications, Deep Learning algorithms that use large amounts of data have become the standard approach \cite{Barsoum2016}. This has led to the creation of large public datasets and to the decoupling of the dataset creation and model training phases. These datasets, despite their usefulness, many times exhibit heavy biases \cite{Prabhu2020} that are easy to overlook for the teams using them. These dataset biases can be found in the demographic proportions of the datasets \cite{Karkkainen2021}, in the relationships between data of multimodal datasets \cite{Zhao2017,Birhane2021}, in the sample labeling and the label themselves \cite{Prabhu2020}, and even in the images of the dataset \cite{Wang2019}. A specific type of bias, the topic of this work, is stereotypical bias \cite{Abbasi2019}, where demographic groups can be equally represented but over or underrepresented in certain categories.

The impact of these biases on the predictions of the final model is highly variable, depending on both the severity and nature of the biases and the context of the application itself. For applications that involve human users, especially when the implementation of the system regulates access to resources or involves the representation of people, unfair predictions can directly lead to harm to population groups \cite{Keyes2018,Avella2020} (allocative and representational harms).

A current area of interest in AI is Facial Expression Recognition (FER) \cite{Li2020}. FER refers to a modality of automatic emotion recognition in which, from a picture of a face, the system predicts the emotional state of the subject. The readiness for implementation, possible with minimal hardware, combined with the nature of the data involved, makes FER an application where biases are easily developed and could potentially lead to representational harms. Furthermore, the face images have some demographic information naturally integrated into them, such as apparent age, gender, or race. With most datasets lacking explicit external demographic labels, bias mitigation techniques are hard to apply, and even bias detection poses a challenge. Regarding gender in particular, although public FER datasets are usually globally balanced, with similar proportions of male and female presenting people, they often hide stereotypical biases. That is, they are unbalanced for certain categories, despite the global balance, which can systematically skew the final model predictions depending on the subject's apparent gender.

In this work, we analyze stereotypical gender bias in the context of FER. In particular, we focus on the FER+ dataset \cite{Barsoum2016}, a refined version of the popular FER2013 dataset \cite{Goodfellow2013}. With this dataset as the base, in our experiments we generate derivative datasets with different amounts of stereotypical bias by altering the gender proportions of certain labels and measuring the variations in the final model predictions. These induced biases allow us to quantify the limits of the variations under extreme stereotypical bias conditions in FER problems. Although previous work \cite{Ahmad2022,Xu2020,Domnich2021,Deuschel2021,Dominguez-Catena2022} has also studied and revealed biases in general and gender biases in particular in FER, no other work has focused on the problem of stereotypical biases in this context. We hope that our contribution will help establish the importance of this type of bias and understand the extent of its impact in this context. 

From our results, we observe a discrepancy in the recognition of certain emotions between genders of up to $29 \%$ under the worst bias conditions. Our results also suggest a safety range for stereotypical bias in the dataset that does not appear to produce bias in the final model. These findings can help future implementations avoid some potential harms in FER due to misrepresentation of groups.

The remainder of this work is organized as follows. Section \ref{sec:related} describes the related work and some background information for our proposal. Next, Section \ref{sec:methodology} describes the proposed experiments and the relevant implementation details. In Section \ref{sec:results} presents and analyzes the results of the experiments. Finally, Section \ref{sec:conclusion} concludes this work and proposes future work.

\section{Related work}\label{sec:related}

\subsection{Facial Expression Recognition}

FER is one of the simplest and most widespread modalities of the more general automatic emotion recognition. In automatic emotion recognition, the system tries to identify the emotional state of a person from their expressions and physiology. Several modalities are possible, depending on both the input data required by the system and the output codification of the emotional state~\cite{Assuncao2022}. FER, in particular, uses as input data a static image or a video of a human face, making it relatively easy to deploy with minimal hardware.

Regarding the emotion codification, the classical approaches are continuous \cite{Mehrabian1996} and discrete models of emotion \cite{Ekman1971}. The continuous model separates emotion into several independent dimensions, such as \textit{valence} and \textit{arousal}. Instead, the discrete model assimilates emotions into several prototypes, with the most common categorization being the six basic emotions of Ekman \cite{Ekman1971}: angry, disgust, fear, happiness, sadness, and surprise. Although the continuous codification is more expressive, the labeling of samples is more subjective and complex. Thus, most FER datasets are based on the discrete approach. In this work, we will focus on the same discrete approach.

\subsection{Bias}

Most definitions of fairness are based on the idea of absence of unwanted bias \cite{Verma2018}. This unwanted bias, understood as a systematic variation in the treatment of a demographic group that can potentially lead to harm.

Although most definitions of bias as a proxy for fairness are designed around the predictions of a model, the general concept of bias can be linked to different sources of bias at different points of the ML life cycle \cite{Suresh2021}. In particular, for applications in ML where public datasets are common, bias present in the source data is particularly relevant \cite{Ntoutsi2020}. Large datasets, in particular, have been subject to extensive analysis, finding different types of bias \cite{Prabhu2020,Denton2021}.

While data bias is predominantly studied in the form of representational bias, where certain demographic groups are overly prevalent in a dataset, another common bias in some types of datasets is \textit{stereotypical bias} \cite{Abbasi2019,Bordalo2016}. In classification tasks, this kind of bias is modeled as a correlation between the demographic attributes of a subject and the problem classes, and can easily leak into the datasets as different demographic profiles for certain classes.

Some works have already analyzed FER systems, finding demographic bias in general \cite{Kim2021,Jannat2021}, including several instances of gender biases \cite{Ahmad2022,Domnich2021,Deuschel2021,Dominguez-Catena2022}. In particular, Ahmad et al. \cite{Ahmad2022} analyzes the prediction of commercial systems, without working with the bias in the original datasets. Domnich and Anbarjafari \cite{Domnich2021} study the gender bias exhibited by six different neural networks trained for FER. Deuschel et al. \cite{Deuschel2021} employ intentionally biased datasets, composed only of male or female subjects, to study the impact of these biases on the detection of action units, a problem closely related to FER. Finally, Dominguez et al. \cite{Dominguez-Catena2022} also uses intentionally biased and balanced datasets to validate a set of metrics for bias detection, using FER as a case study and showing inherent representational and stereotypical biases in some FER datasets.

Unlike the previous work, we will focus on the stereotypical bias in FER. We will employ progressively biased datasets to measure the impact of this type of bias on the trained model. Bias is often measured with specific bias metrics, which helps quantify its impact. Despite this, it is important to notice that any application of a specific metric still requires a proper qualitative discussion of its context, or it can easily lose its usefulness \cite{Schwartz2022}. For this reason, in this work, we will employ a qualitative and intuitive approach without employing a specific metric. Nonetheless, we will look for deviations in recalls (accuracy constrained to the examples of a certain class) between demographic groups, with an underlying notion of fairness consistent with the \textit{conditional use accuracy equality} \cite{Berk2018}. To the best of our knowledge, no other work on FER has focused on this type of bias, and most have only focused on representational bias.

\section{Methodology}\label{sec:methodology}

\subsection{Datasets}

In this work, we employ the FER+ dataset\cite{Goodfellow2013}, based on FER2013 \cite{Goodfellow2013}. FER2013 \cite{Goodfellow2013} is one of the most popular publicly available \textit{in the wild} FER datasets, with more than $32,000$ labeled images obtained from Internet searches. The images in the original dataset were automatically annotated, leading to systematic inaccuracies, which were later corrected by FER+ \cite{Barsoum2016}, a relabeling of the same image set. The images in FER+ are grayscale and have a small resolution of $48 \times 48$ pixels. This small image size supports fast and resource-light model training, one of the main reasons for its popularity.

\subsection{Demographic Relabeling} \label{ssec:demrelabel}

As FER2013 is not gender labelled, we use an external model, FairFace \cite{Karkkainen2021} to obtain an apparent gender prediction for each image. The FairFace model was trained on the homonymous dataset, composed of $108,501$ images labeled for apparent gender, apparent race, and apparent age. In the original experiments, the model achieved an accuracy greater than $92\%$ for gender recognition in FairFace and three other demographic datasets. The model is publicly available\footnote{\url{https://github.com/joojs/fairface}}.

It is important to note that FairFace comes with some serious limitations. Although this is particularly evident in the race categories, limited to six stereotypical groups, namely White, Black, East Asian, Southeast Asian, Latino, Indian, and Middle Eastern, it is also present in the gender category. For the creation of FairFace, as is still common for most gender-labeled datasets, external annotators manually labeled gender into a binary classification of \emph{Male} and \emph{Female}. This classification correlates with how many societies identify gender, but can easily misrepresent people, as is the case for binary and non-binary transgender people and other gender non-conforming individuals \cite{Keyes2018}. Nevertheless, as almost no datasets have the required demographic information, proxy labels such as the ones provided by FairFace give us a reasonable overview of the population of the datasets, even if they could be unreliable for the individual subjects. Additionally, as the real demographic information is also unknown to the trained FER models, if bias is present in them it must be based only on the physical appearance. Thus, we perform our analysis on these labels, as they can help uncover biases based on these apparent demographic characteristics, even if they do not always correlate with the true self-reported characteristics. Any bias based on the apparent characteristics predicted by the auxiliary model must be considered under these limitations, and further work must be done to test if the bias is still present when we consider the real demographic characteristics.

\subsection{Generation of derivative datasets}\label{ssec:derdatasets}

To study the impact of stereotypical bias, we generate three types of datasets, namely, stratified, balanced, and biased. All of these are created as subsets from the original FER+ dataset.

\subsubsection{Stratified subsets.}

{
    \def\OldComma{,}
    \catcode`\,=13
    \def,{%
      \ifmmode%
        \OldComma\discretionary{}{}{}%
      \else%
        \OldComma%
      \fi%
    }%
To enable the comparison between different datasets, we implement a method to generate stratified subsets from a source dataset with a given target size, expressed as a ratio $r \in [0, 1]$ of the number of examples in the original dataset. To generate a stratified version, we consider both the set of target classification labels $L = \{\text{angry}, \text{disgust}, \text{fear}, \text{happy}, \text{sad}, \text{surprise}, \text{neutral}\}$, and the demographic groups of interest, in our case $S = \{\text{male}, \text{female}\}$ and their combinations defined by the Cartesian product $L \times S$. For each of these combinations independently, we perform a random subsample with target ratio $r$. This process guarantees that the relative proportions between each label in $L$, the demographic group in $S$, or the combination of both in $L \times S$ are kept, while the overall size is reduced by the desired ratio $r$ (plus or minus some rounding error). Thus, the stratified datasets maintain the same stereotypical deviations and general demographic proportions as the source data set.}

\subsubsection{Balanced subsets.}

As the original FER+ dataset already contains some stereotypical bias \cite{Dominguez-Catena2022}, we generate a balanced version of it to serve as a general baseline. This dataset has the same proportions of each label in $L$ as the original FER+, but for each of them, the proportions of the demographic groups in $S$ are equalized. To generate this balanced dataset, we first calculate the most underrepresented group $(l, s) \in L \times S$ by calculating the imbalance ratio of each one in their respective label $l$:
\begin{equation}
    \text{imb}(l, s) = \frac{|\{x | x \in D_l \opand x \in D_s \}|}
                            {|\{x | x \in D_l                \}|}\ ,
\end{equation}
where $D_l$ denotes the subset of the dataset samples labeled with $l$ and $D_s$ the subset identified as part of the demographic group $s$.

After this, we subsample each of the groups independently according to:
\begin{equation}
    \text{ratio}(l, s) = \frac{\min_{l'\in L, s' \in S}{\text{imb}(l', s')}}{\text{imb}(l,s)}\ .
\end{equation}

The resulting dataset keeps the distribution of the target labels while making the demographic groups in each label and in the whole dataset equally represented.

\subsubsection{Biased subsets.}

Finally, we also generate intentionally stereotypically biased datasets. These datasets are built from the balanced datasets, but inducing a certain amount of bias into one of the labels $l$ with respect to a target demographic group $s$. The amount of induced bias $b \in [-1, 1]$ is applied as:

\begin{itemize}
    \item If $b < 0$, a negative bias is introduced, that is, the target demographic group $\{x | x \in D_l \opand x \in D_s\}$ is reduced by the ratio $1 + b$.  The examples labeled as $l$ belonging to the other demographic groups are kept intact.
    \item If $b > 0$, a positive bias is introduced, that is, the target demographic group is left intact, reducing the representation of the rest of the samples $\{x | x \in D_l \opand x \notin D_s\}$ by the ratio $1-b$.
    \item If $b = 0$, the balanced dataset is not modified, and no bias is introduced.
\end{itemize}

After biasing the target label $l$, the resulting number of examples of that label is $1 - \frac{|b|}{2}$ of the original label support. To compensate for this effect, the other labels are also subsampled by the ratio $1 - \frac{|b|}{2}$. This reduces the final dataset size, but keeps the label distribution equal to the original dataset.

The resulting dataset has, for $b = -1$ a total absence of the target demographic group in the label (underrepresentation), for $b = 1$ only samples of the target demographic group in the label (overrepresentation), and for $b = 0$ is balanced. The intermediate values allow for fine control of the amount of bias. In all cases, the label distribution is kept identical to the original dataset.

\subsection{Experiments}

In our experiments, we aim to generate biased datasets in the extremes of the stereotypical bias possibilities and then measure the final model accuracy imbalances for the relevant demographic groups and labels. For this, we first obtain the demographic profile of FER+ in the gender category. With this information, we chose some of the more heavily biased labels and generate datasets that exaggerate those same biases. The biased datasets are generated with different degrees of bias, from a negative bias of $-1$ to a positive one of $1$ in steps of $0.2$, all of them with respect to the "female" class as recognized by FairFace. The balanced datasets will serve as a baseline, showing the behavior expected in the absence of stereotypical bias for a certain dataset size. 

To analyze the influence of the datasets on the performance of the model, we train a model for each generated dataset and obtain the predictions over the whole FER+ test partition. We then obtain the recall for each combination of dataset, label, and gender group, that is, the accuracy of the classifier for the examples belonging to the specific gender group and with a certain true label. In particular, we expect to obtain the maximum difference in recall between the demographic categories \emph{male} and \emph{female} in the extreme biased datasets for the biased labels, as a measure of the maximum impact of stereotypical bias on the recognition of the affected labels.

\subsection{Experimental Setup}

We employ a simple VGG11 \cite{Simonyan2015} network with no pretraining as the base test model. This is a classical convolutional architecture often used as a baseline for machine learning applications. The experiments are developed on PyTorch 1.10.0 and Fastai 2.6.3. The hardware used is a machine equipped with a GeForce RTX 2060 Super GPU, 20~GB of RAM, an Intel\textregistered\ Xeon\textregistered\ i5-8500 CPU, and running Ubuntu Linux 20.04.

All the models are trained under the same conditions and hyperparameters, namely, a maximum learning rate of $1e^{-2}$ with a 1cycle policy (as described in \cite{Smith2018} and implemented in Fastai) for $20$ iterations. This parameter was decided using the \textit{lr\_finder} tool in Fastai. The batch size is set to $256$, the maximum allowed by the hardware setup. For each dataset, we train the model $10$ times and average the results over them. We have also applied the basic data augmentation provided by Fastai through the \textit{aug\_transforms} method, including left-right flipping, warping, rotation, zoom, brightness, and contrast alterations.

For each dataset configuration to be tested, we perform ten individual training processes, for each one regenerating a new resampled dataset to ensure that the sampling process does not affect the final results.

\section{Results and Discussion}\label{sec:results}

\subsection{Dataset initial bias}

\begin{figure}
    \centering
    \resizebox{\columnwidth}{!}{
    \input{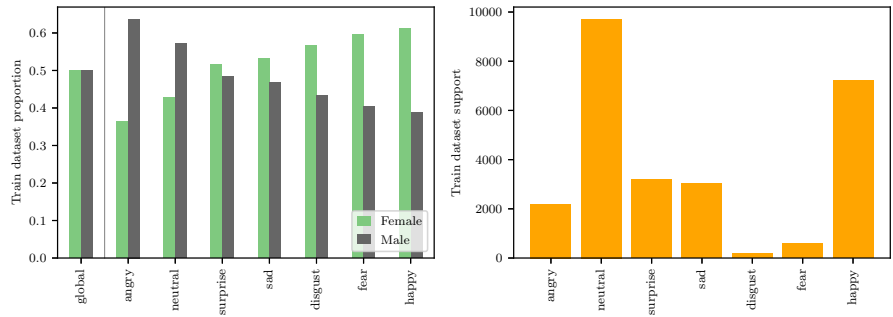}}
    \caption{FER+ gender distribution and support by label.}
    \label{fig:fig_genderdistrib}
\end{figure}

We perform the demographic relabeling of FER+ with the FairFace public model, as described in Section \ref{ssec:demrelabel}. The proportions of the gender category in the whole dataset and for each label are shown in Figure \ref{fig:fig_genderdistrib}, together with the label supports. The global gender proportions are almost uniform, at $50.1\%$ for the \emph{Female} group and $49.9\%$ for the \emph{Male} group, showing very little direct representational bias. For stereotypical bias, the individual labels show a much greater disparity. The two extremes are the label \emph{angry}, with an underrepresentation of the \emph{Female} group ($36.27\%$ of the label support) and the label \emph{happy}, with an underrepresentation of the \emph{Male} group ($38.7\%$ of the label support). The rest of the labels in the dataset lie in between, with slightly lower imbalances.

Interestingly, the biases found in the labels \emph{happy} and \emph{angry} are consistent with the classical \textit{angry-men-happy-women} bias, a psychological bias pattern well researched in the expression and recognition of human emotions \cite{Kring1998,Atkinson2005}.

\subsection{Induced bias impact}

\begin{figure}
    \centering
    \resizebox{\columnwidth}{!}{
    \input{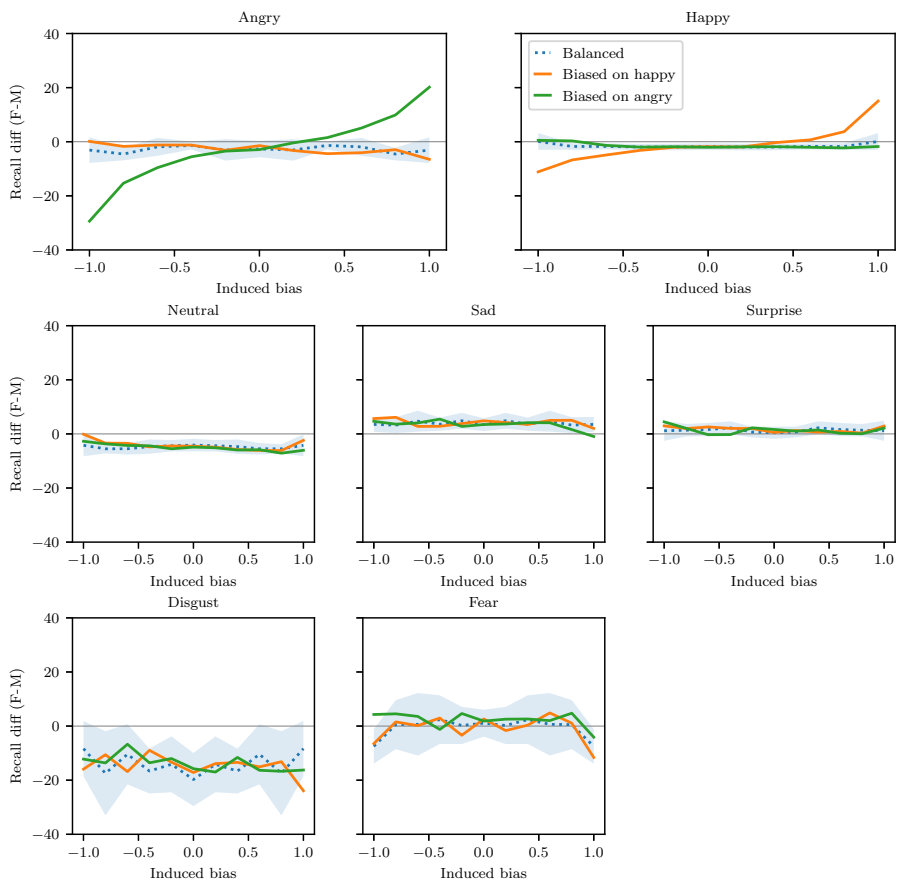}}
    \caption{Recall difference Male-Female in the different emotion labels. Positive numbers mean a higher recall for the \textit{Female} group than for the \textit{Male} one. The baseline balanced datasets are plotted according to size, aligned with the corresponding biased datasets.}\label{figure:recalldiff}
\end{figure}

The recall results obtained by the models are shown in Table \ref{tab:results_by_gender}. For brevity, only the results for the four most extremely biased datasets and the size-equivalent stratified balanced dataset are reported in the Table, with the complete results being graphically presented in Figure \ref{figure:recalldiff}. The difference between gender recalls is highest for the biased datasets in all cases, with the largest absolute difference found in the labels \textit{angry}, \textit{disgust}, and \textit{happy}. In particular, for the \textit{angry} label, biasing against the \textit{Female} group maximizes the recall difference at $29.36 \%$ in favor of the \textit{Male} group, while for the \textit{happy} label it is the positive bias in favor of the \textit{Female} group that maximizes the difference at $15.03 \%$. Biasing the \textit{angry} label generates a total range of disparity of $49.53 \%$ between its extremes, while biasing the \textit{happy} label of $26.17 \%$. 

The label \textit{disgust} seems to be a particular case, with the largest recall differences between the gender groups overall. Difference values range from a $8.33 \%$ difference to a $23.92 \%$ difference, always in favor of the \textit{Male} group. Recall that none of the biased datasets are designed to bias in this label, that is kept balanced in all the derivative datasets, and even in the original dataset exhibits only a mild stereotypical bias against the \textit{Male} group, which constitutes a $43.25 \%$ of the original support. However, this label also has the lowest support in the original FER+ dataset, with the lowest general recalls of all labels for all configurations. The label \textit{disgust} also shows the highest standard deviation, between $\pm 10.22$ and $\pm 16.83$, making the results for this label unreliable.

The rest of the labels show some variations in general, but generally seem unaffected by the bias induced in \textit{angry} and \textit{happy}. An exception seems to be in the application of the positive bias in the \textit{happy} label, overrepresenting the \textit{female} group for that label, which seems to decrease the recall for the \textit{angry} label of the same group.

\begin{table}[htb!]
\resizebox{\columnwidth}{!}{
\begin{tabular}{rrrrrm{2em}rrr}
\toprule
         &      &  \multicolumn{3}{c}{Happy}  & & \multicolumn{3}{c}{Angry}  \\
\cline{3-5}\cline{7-9}
         &      &  Female $-1.00$ &      Female $0.00$ &   Female $+1.00$ & & Female $-1.00$ &      Female $0.00$ & Female $+1.00$ \\
\midrule
{} & Size &              $9475$ &             $9476$ &              $9475$ & &             $9477$ &             $9476$ &              $9477$ \\
\hline
\cc angry & \cc Male &    $74.40 \pm 1.47$ &   $76.30 \pm 3.40$ &    $74.24 \pm 4.73$ & & \cc   \pmb{$81.25 \pm 2.61$} & \cc   $76.30 \pm 3.40$ &  \cc   $55.87 \pm 2.24$ \\
\cc         & \cc Female &    $74.53 \pm 3.45$ &   $73.21 \pm 3.33$ &    $67.74 \pm 3.34$ &  & \cc   $51.89 \pm 3.77$ &  \cc  $73.21 \pm 3.33$ &  \cc   \pmb{$76.04 \pm 1.99$} \\
\cc         & \cc Diff &     $0.13 \pm 3.75$ &   $-3.10 \pm 4.76$ &    $-6.50 \pm 5.79$ &  & \cc  \pmb{$-29.36 \pm 4.59$} & \cc   $-3.10 \pm 4.76$ &  \cc   $20.17 \pm 2.99$ \\
\hline
neutral & Male &    $66.45 \pm 3.10$ &   $71.52 \pm 2.71$ &    $67.24 \pm 4.66$ & &    $70.72 \pm 1.84$ &   $71.52 \pm 2.71$ &    \pmb{$74.91 \pm 1.66$} \\
         & Female &    $66.33 \pm 2.77$ &   $67.29 \pm 3.00$ &    $64.86 \pm 3.79$ & &    $67.99 \pm 2.48$ &   $67.29 \pm 3.00$ &    \pmb{$68.85 \pm 1.26$} \\
         & Diff &    $-0.12 \pm 4.15$ &   $-4.22 \pm 4.04$ &    $-2.38 \pm 6.01$ & &    $-2.73 \pm 3.09$ &   $-4.22 \pm 4.04$ &    \pmb{$-6.06 \pm 2.09$} \\
\hline
surprise & Male &    $80.50 \pm 3.91$ &   $83.71 \pm 2.58$ &    $81.93 \pm 3.16$ & &    $82.77 \pm 2.32$ &   $83.71 \pm 2.58$ &    \pmb{$84.16 \pm 1.71$} \\
         & Female &    $83.46 \pm 3.06$ &   $84.88 \pm 2.67$ &    $84.83 \pm 2.51$ & &    \pmb{$87.22 \pm 2.75$} &   $84.88 \pm 2.67$ &    $86.29 \pm 1.87$ \\
         & Diff &     $2.97 \pm 4.97$ &    $1.17 \pm 3.71$ &     $2.90 \pm 4.03$ & &     \pmb{$4.45 \pm 3.60$} &    $1.17 \pm 3.71$ &     $2.13 \pm 2.54$ \\
\hline
sad & Male &    $62.22 \pm 3.71$ &   $66.82 \pm 1.53$ &    $66.88 \pm 3.90$ & &    \pmb{$69.66 \pm 2.53$} &   $66.82 \pm 1.53$ &    $69.55 \pm 2.69$ \\
         & Female &    $67.88 \pm 2.71$ &   $70.33 \pm 2.30$ &    $68.91 \pm 3.26$ & &    \pmb{$74.29 \pm 2.25$} &   $70.33 \pm 2.30$ &    $68.59 \pm 2.72$ \\
         & Diff &     \pmb{$5.66 \pm 4.60$} &    $3.51 \pm 2.76$ &     $2.04 \pm 5.09$ & &     $4.63 \pm 3.39$ &    $3.51 \pm 2.76$ &    $-0.96 \pm 3.82$ \\
\hline
disgust & Male &   $51.25 \pm 16.01$ &   $45.00 \pm 9.19$ &   \pmb{$61.25 \pm 11.46$} & &    $56.88 \pm 9.86$ &   $45.00 \pm 9.19$ &   $56.25 \pm 12.18$ \\
         & Female &    $35.33 \pm 5.21$ &   $36.67 \pm 4.47$ &    $37.33 \pm 6.11$ &  &   \pmb{$44.67 \pm 7.33$} &   $36.67 \pm 4.47$ &    $40.00 \pm 7.30$ \\
         & Diff &  $-15.92 \pm 16.83$ &  $-8.33 \pm 10.22$ &  \pmb{$-23.92 \pm 12.98$} & &  $-12.21 \pm 12.29$ &  $-8.33 \pm 10.22$ &  $-16.25 \pm 14.20$ \\
\hline
fear & Male &    $66.25 \pm 5.73$ &   $65.83 \pm 4.86$ &    \pmb{$68.75 \pm 4.66$} & &    $56.67 \pm 7.73$ &   $65.83 \pm 4.86$ &    $59.58 \pm 7.23$ \\
         & Female &    $59.76 \pm 3.76$ &   $58.33 \pm 4.16$ &    $57.14 \pm 5.11$ & &    \pmb{$60.95 \pm 5.02$} &   $58.33 \pm 4.16$ &    $55.48 \pm 6.21$ \\
         & Diff &    $-6.49 \pm 6.85$ &   $-7.50 \pm 6.40$ &   \pmb{$-11.61 \pm 6.91$} & &     $4.29 \pm 9.21$ &   $-7.50 \pm 6.40$ &    $-4.11 \pm 9.53$ \\
\hline 
\cc happy & \cc Male &    \cc $89.10 \pm 2.46$ &   \cc $87.66 \pm 2.23$ &    \cc $72.91 \pm 3.47$ & &    $87.90 \pm 2.11$ &   $87.66 \pm 2.23$ &    \pmb{$89.79 \pm 1.71$} \\
\cc         & \cc Female &    \cc $77.96 \pm 1.86$ &   \cc $87.75 \pm 2.14$ &    \cc $87.94 \pm 1.87$ & &    \pmb{$88.42 \pm 1.89$} &   $87.75 \pm 2.14$ &    $88.02 \pm 1.48$ \\
\cc         & \cc Diff &   \cc $-11.14 \pm 3.09$ &    \cc $0.09 \pm 3.09$ &    \cc \pmb{$15.03 \pm 3.94$} & &     $0.52 \pm 2.83$ &    $0.09 \pm 3.09$ &    $-1.77 \pm 2.26$ \\
\bottomrule
\end{tabular}
}
\caption{Recall by label and gender for the key datasets analyzed. For the difference between gender recalls, the highest absolute value is in bold.}\label{tab:results_by_gender}
\end{table}

The difference between the recalls of the \textit{male} and \textit{female} groups for each degree of induced bias is shown in Figure \ref{figure:recalldiff}. In the Figure, the vertical axis corresponds to the difference in recall from the \textit{female} group to the \textit{male} group, and the horizontal axis corresponds to the amount of induced bias. The difference of recall obtained in the balanced datasets is included as the comparison baseline.

For all labels, if no bias is introduced on that particular label, the recall differences are close to the baseline levels. When observing the differences in the recall of the affected label for the biased datasets, the effect of the dataset bias becomes apparent. For both the \textit{angry} and \textit{happy} labels, the negative biases, which correspond to an under-representation of the \textit{female} group on the label, show a difference in recall in favor of the \textit{male} group, and the opposite is observed for positive amounts of bias. For the datasets biased in the \textit{angry} label with a negative or positive bias for the \textit{female} group, the difference in recalls of the \textit{angry} label quickly deviates from baseline levels when exposed to a bias of $\pm 0.2$ or greater, exceeding $\pm 20\%$ of difference under extreme bias conditions. For the label \textit{happy}, the effect is not as pronounced and only when trained on datasets with bias of $\pm 0.4$ or higher does the difference in recalls deviate from the baseline behavior.



For both the labels \textit{angry} and \textit{happy}, a safe zone can be observed where bias in the dataset does not significantly affect the difference in recalls. The behavior of the biased model when trained under this limited amount of bias seems to be similar to the baseline dataset. In the case of the label \textit{happy}, this safe zone includes the datasets with a stereotypical bias of $\pm 0.4$ and lower, while on \textit{angry} it is more restricted, including only those with a stereotypical bias of $\pm 0.2$ and lower.

\section{Conclusion}\label{sec:conclusion}

In this work, we have studied the impact of stereotypical bias in FER datasets and their resulting models through the induction of controlled bias in the dataset. In particular, for the FER problem, we have observed up to a $29 \%$ disparity in the recognition of certain emotions, namely \textit{angry}, when the dataset lacks representation of a gender category for the label. We have shown that this kind of bias is already present in publicly available datasets, in particular in FER+, but our experiments suggest that a small amount of stereotypical bias in the gender category seems acceptable, not impacting the final performance for the underrepresented group. Nevertheless, it is important to notice that the acceptable amount of stereotypical bias seems to be context-dependent, varying at least between labels. Our findings support the importance of a thorough bias analysis of public datasets in problems like FER, where a global balance of demographic representation in the dataset can still hide other types of bias that harm certain demographic groups.

In light of our findings, we highly recommend that future datasets, especially those created from Internet searches and intended for public release, are tested for stereotypical bias and corrected accordingly by down-sampling the overrepresented demographic groups. Although other mitigation techniques could be performed later in the training phases, this type of bias is easy to overlook and can leak into bias in the trained models if left untreated. Furthermore, we strongly advise dataset creators to include the relevant demographic information of the subjects when possible, to allow the future study of new forms of demographic bias in their datasets.

A problem that requires further analysis is the large differences in the gender recall of certain labels, such as \textit{disgust}. This difference is present even for the balanced versions of the dataset, suggesting a measurement bias or an inherent representation problem in this label. The label \textit{disgust}, in particular, has low support, which could imply that stereotypical bias problems have a greater impact in smaller datasets. Further work is also required to replicate these results for other datasets, models and different applications. The development of properly labeled datasets that include demographic information of the represented subjects would also solidify this analysis, currently limited by the demographic relabeling model employed.

\section*{Acknowledgments}
This work was funded by a predoctoral fellowship of the Research Service of Universidad Publica de Navarra, the Spanish MICIN (PID2019-108392GB-I00 and PID2020-118014RB-I00 / AEI / 10.13039/501100011033), and the Government of Navarre (0011-1411-2020-000079 - Emotional Films).

\bibliographystyle{splncs04}
\bibliography{FER_bibtex}

\end{document}